\documentclass[journal]{IEEEtran}
\usepackage[colorinlistoftodos]{todonotes}
\usepackage{cite}
\usepackage{booktabs}
\usepackage{amsthm}
\usepackage{newtxmath}

\usepackage{multirow}
\usepackage{subfigure}
\usepackage{graphicx}
\usepackage[ruled,vlined]{algorithm2e}
\usepackage[utf8]{inputenc} 
\usepackage[T1]{fontenc}    
\usepackage{url}            
\usepackage{amsfonts}       
\usepackage{nicefrac}       
\usepackage{microtype}      
\usepackage{xcolor}         
\usepackage{graphicx}
\usepackage{multirow}
\usepackage{caption}

\begin{document}

\title{Boundary-aware Graph Reasoning \\ for Semantic Segmentation}

\author{Haoteng~Tang$^{*}$, Haozhe~Jia$^{*}$, Weidong~Cai, Heng~Huang, Yong~Xia$^{\dag}$, Liang~Zhan$^{\dag}$

\thanks{H.Tang, H.Jia, H.Huang, L.Zhan are with the Department
of Electrical and Computer Engineering, University of Pittsburgh}
\thanks{H.Jia, Y.Xia are with the School of Computer Science and Engineering, Northwestern Polytechnical University}
\thanks{W. Cai is with the School of Computer Science, University of Sydney}

\thanks{$^{*}$ are co-first authors.}
\thanks{$^{\dag}$ are co-corresponding authors.}
}

\markboth{Journal of \LaTeX\ Class Files,~Vol.~XX, No.~XX, August~2021}%
{Shell \MakeLowercase{\textit{et al.}}: Bare Demo of IEEEtran.cls for IEEE Journals}

\maketitle

\begin{abstract}
In this paper, we propose a Boundary-aware Graph Reasoning (BGR) module to learn long-range contextual features for semantic segmentation.
Rather than directly construct the graph based on the backbone features, our BGR module explores a reasonable way to combine segmentation erroneous regions with the graph construction scenario. 
Motivated by the fact that most hard-to-segment pixels broadly distribute on boundary regions, our BGR module uses the boundary score map as prior knowledge to intensify the graph node connections and thereby guide the graph reasoning focus on boundary regions.
In addition, we employ an efficient graph convolution implementation to reduce the computational cost, which benefits the integration of our BGR module into current segmentation backbones. 
Extensive experiments on three challenging segmentation benchmarks demonstrate the effectiveness of our proposed BGR module for semantic segmentation.
\end{abstract}


\IEEEpeerreviewmaketitle

\section{Introduction}

\IEEEPARstart{R}{ecently}, state-of-the-art methods based on fully convolutional networks (FCNs) \cite{long2015fully} have made tremendous progress in semantic segmentation and have shown convincing effectiveness in dense pixel prediction tasks.
However, it is well recognized that the segmentation performance is still constrained by its insufficient capability of reasoning long-range contextual information, since convolutional operations process visual features with a limited receptive field and thus can only capture local information.

Several methods have been proposed to address this issue from different perspectives.
Among them, two popular solutions are (1) utilizing the dilated convolution, large kernel convolution, and spatial pyramid pooling to enlarge the receptive field and learn multi-scale context \cite{chen2018deeplab,chen2017rethinking,peng2017large,zhao2017pyramid,chen2018encoder} and (2) building different self-attention modules to capture and aggregate global information from all locations for each pixel \cite{wang2018non,fu2019dual,chen20182,zhao2018psanet,li2019expectation,huang2019ccnet}.
In these methods, the global context learning is nevertheless still limited, since it only performs low-level long-range reasoning and relies on inefficient computation in practical use.
Graph-based models have been introduced as feasible solutions to reasoning global relationships for semantic segmentation due to its inherent information propagation ability \cite{chen2019graph,li2020spatial,liang2018symbolic}. These models, however, perform graph reasoning without using any prior knowledge. For instance, it would be more efficient for a graph model to reason long-range dependencies if it focuses more on the regions with segmentation errors, since most pixels can be correctly classified by the segmentation backbone.

In this paper, we explore how to introduce erroneously segmented regions as prior knowledge to the graph reasoning, and thus to boost its discriminatory power.
Considering that most segmentation errors occur in boundary regions \cite{yuan2020segfix}, boundary regions can serve as a natural substitution of erroneously segmented regions. 
Based on this, we propose a Boundary-aware Graph Reasoning (BGR) module which strengthens the graph reasoning in boundary regions.
Specifically, based upon the visual features produced by the backbone, the BGR module first constructs a graph using a pixel-to-node mapping strategy.
Then, a boundary score map is generated to re-weight graph node connections so as to intensify the boundary information, which is named as the `boundary-aware' operation. 
Finally, the graph convolution-based global reasoning is performed to form the long-range context enhanced features, which are fused with the backbone features for the final prediction.

The main contributions of our work can be summarized as follows.
(1) The proposed BGR module utilizes graph reasoning to capture long-range dependencies for segmentation models. The boundary discrimination is further introduced to the BGR module, enabling hard-to-segment pixels to learn better global contextual representation during the reasoning.
To the best of our knowledge, this is the first to use boundary prior knowledge to facilitate graph reasoning for semantic segmentation.
(2) An efficient implementation of graph convolutions is developed to perform graph reasoning with a significantly reduced computational cost. As a result, our BGR module can be easily incorporated into any existing segmentation backbones.
(3) We evaluated the BGR module extensively on three image segmentation benchmarks (\textit{i.e.}, PASCAL VOC, COCO-Stuff, and Cityscapes), and our results suggest that using the boundary prior knowledge can improve the reasoning for global context and the proposed BGR module can boost the segmentation backbone to achieve the state-of-the-art performance.

\section{Related Work}
\subsection{Global Context Learning for Semantic Segmentation}
In recent years, many FCN variants have been proposed to reduce the limitations of local convolutional operations.
To build long-range dependencies, the conditional random field (CRF) \cite{chandra2017dense} has been introduce to FCNs \cite{zheng2015conditional,chen2018deeplab}.
PSP-Net \cite{zhao2017pyramid} and DeepLab models \cite{chen2017rethinking,chen2018encoder,chen2018deeplab} utilize multi-scale dilated convolutions and spatial pyramid pooling to enlarge the receptive field of traditional convolutional operations and aggregate multi-scale contextual information.
GCN \cite{peng2017large} combines convolution decomposition and large kernel convolution to enlarge the valid receptive field and thereby learns better long-range context.
Besides, other attempts \cite{wang2018non,fu2019dual,chen20182,zhao2018psanet,li2019expectation,huang2019ccnet} introduce the self-attention mechanism \cite{vaswani2017att} to construct global contextual representations by exploiting the correlations between each pixel and all other pixels in the feature space.
Due to the excessive and indiscriminate fashion of information aggregation, the global context learning in these methods is still limited and inefficient.
Recently, the graph-base reasoning with a superior high-level reasoning ability has been increasingly used for computer vision applications.
In \cite{li2018beyond}, GCU categories all spatial pixels into different regions in the graph space and further propagates information across all region vertices to capture long-range dependencies among these regions.
Besides, GloRe \cite{chen2019graph} and SGR \cite{liang2018symbolic} directly apply feature transformation and convolutional mapping to globally aggregated local features over the coordinate space into an interaction graph space for further relational reasoning.
These methods, however, require a predetermined node number for feature mapping, which tends to reduce their self-adaptability.
Moreover, the spatial relationship among the pixels of visual features can not be maintained, as the graph reasoning is performed based on regions instead of pixels.
CDGCNet \cite{hu2020class} and SpyGR \cite{li2020spatial} directly perform graph reasoning in the feature space through pixel-wise projections. Nevertheless, they either restrict only the graph connections among sampled pixels or perform relational reasoning without using any prior knowledge.
In contrast, our BGR module not only adopts the pixel-to-node mapping to maintain the spatial information among the graph nodes, but also introduces boundary discrimination to intensify the relation reasoning on hard-to-segment regions.

\subsection{Semantic Boundary Learning}
Semantic boundary localization is a fundamental task in computer vision. Many efforts \cite{lin2017refinenet,yu2018learning,takikawa2019gated,ding2019boundary,zhen2020joint,yuan2020segfix} have been devoted to further boost the segmentation performance via exploiting semantic boundaries.
The most typical stream \cite{yu2018learning,ding2019boundary} has a multi-task learning framework, under which an additional boundary detection branch is constructed to facilitate the existing segmentation branch with semantically discriminative features.
The dual loss regularizer is also adopted as a consistency constraint to further improve the segmentation performance \cite{takikawa2019gated,zhen2020joint}. 
In this paper, we do not follow previous approaches to design complex components and special fusion module for the boundary detection branch but merely apply basic linear convolutional layers to generate the boundary score map as the prior knowledge for the subsequent graph reasoning.

\section{Method}
\begin{figure*}[tp]
\centering
\includegraphics[width=1\textwidth]{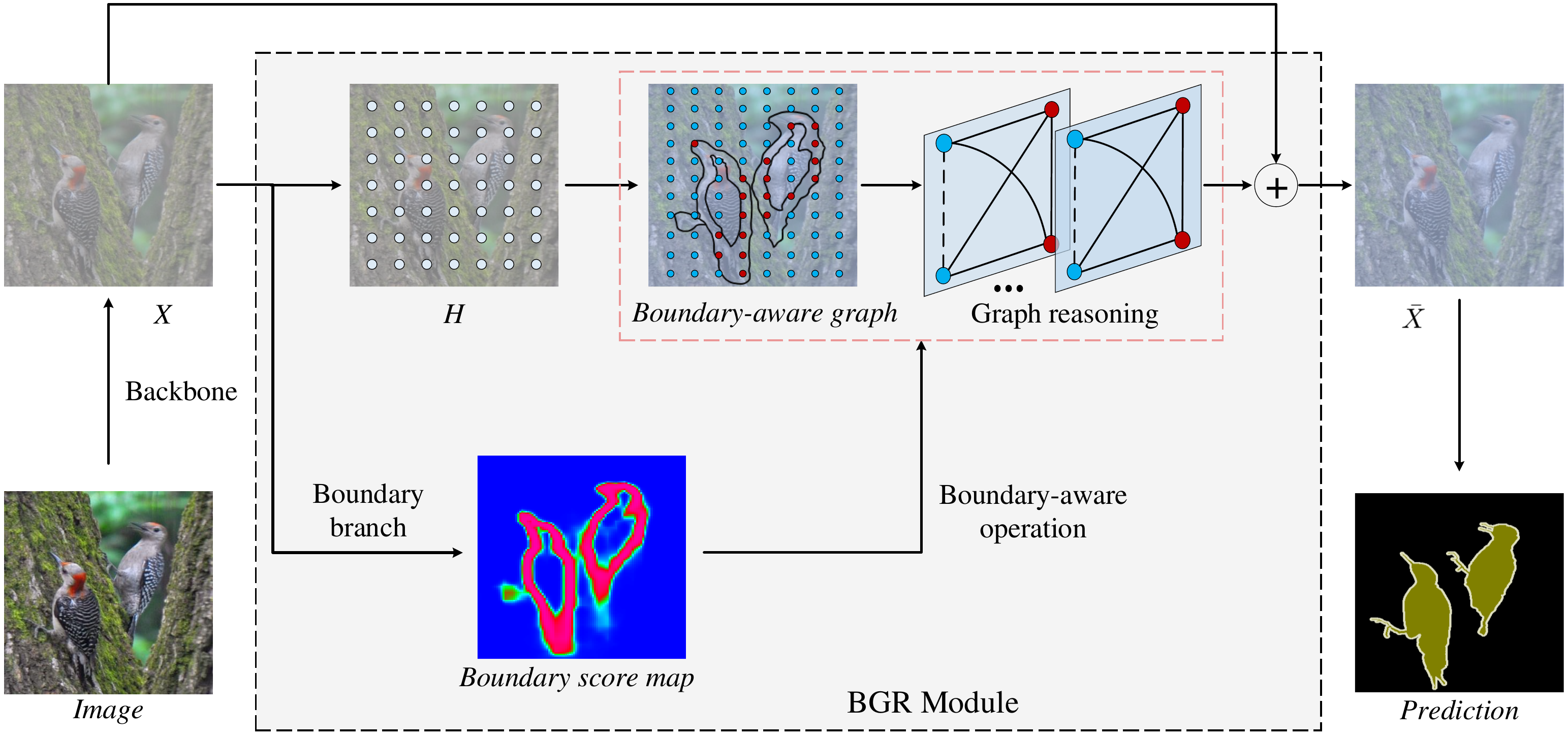}
\caption{An overview of the proposed BGR module. First, a boundary score map is generated to locate the graph nodes on the object boundary region (red nodes in the boundary-aware graph). Then, the node connectivity is re-weighted by the boundary score map so that the connectivity related to the boundary nodes is intensified (solid lines in graph reasoning) comparing to the connectivity among non-boundary nodes (dash lines in graph reasoning). Next, the graph reasoning is performed on the constructed boundary-aware graph to learn long-range contextual information. Finally, the enhanced features are fused with the backbone features for pixel-wise segmentation prediction.}
\label{fig1}
\end{figure*}

In this section, we present our Boundary-aware Graph Reasoning (BGR) module to improve the performance of semantic segmentation models.
Besides performing graph reasoning to capture efficient long-range dependencies, the boundary-aware mechanism in our BGR module can introduce extra semantic discrimination to guide graph reasoning and make it focus on those hard-to-segment pixels.
In addition, we specifically design an efficient implementation of graph convolution to perform global reasoning on the constructed boundary-aware graph. This implementation enables our BGR module to be flexibly embedded into existing segmentation backbones.

The basic idea of our BGR module is illustrated in Figure \ref{fig1}.
For a 2D backbone feature map $X \in \mathcal{R}^{h \times w \times c}$ with $c$ being the channel number and $h, w$ being spatial dimensions, the BGR module first adopts a pixel-wise mapping to project $X$ into the graph space as node features $H \in \mathcal{R}^{N \times c}$, where $N=h\times w$.
The connectivity between each pair of nodes is measured based on the similarity between their features.
Meanwhile, a boundary score map is generated by a boundary detection branch, where the score of each pixel represents the probability of the pixel locating at the boundary region.
To focus graph reasoning on those hard-to-segment pixels, the boundary score map is then introduced as the prior knowledge to the graph, aiming to re-weight the node connectivity so that the connectivity related to boundary nodes can be highlighted.
Next, graph reasoning is performed by means of graph convolution operations, which enable graph nodes to aggregate global context via information propagation.
Due to the designed boundary-aware operation, the reasoning of boundary nodes is particularly intensified.
Finally, the global context aggregated feature is fused with original 2D backbone feature $X$ to form the output feature map $\bar{X}$ of our BGR module, which is then forwarded to the segmentation backbone for the pixel-wise segmentation prediction. We now delve into the details of the proposed BGR module.

\subsection{Boundary-aware Graph Construction}
Given an attribute weighted graph $G=(A, H)$, where $A\in \mathcal{R}^{N\times N}$ is the adjacency matrix of $G$ and $H \in \mathcal{R}^{N \times c}$ is the node feature matrix. 
The convolutional layer on graph $G$ is given as \cite{kipf2016semi}:
\begin{equation}
    H^{(l+1)} = \sigma (\tilde{D}^{-\frac{1}{2}} \tilde{A} \tilde{D}^{-\frac{1}{2}} H^{(l)} \Theta^{(l)}),
\label{eq1}
\end{equation}
where $\tilde{A} = A + I$, $\tilde{D}_{ii} = \sum_{j}\tilde{A}_{ij}$, $H^{(l)}$ is the input node feature matrix of the $l$-th layer, and $\Theta^{l}$ is the trainable parameters in $l$-th layer, and $\sigma(\cdot)$ is a non-linear activation function, $e.g.$, ReLU.
The initial input node feature matrix $H^{(1)}=H$ is the original one.

To accurately represent pixels by graph nodes and maintain the spatial relationships among pixels, we perform the pixel-to-node mapping strategy. 
Given the backbone generated feature map $X \in \mathcal{R}^{h \times w \times c}$, the node feature matrix $H$ can be obtained by:
\begin{equation}
    H = reshape(\phi(X)) \in \mathcal{R}^{N \times c},
\end{equation}
where
$\phi(\cdot)$ is a linear embedding operation (e.g., $1 \times 1$ convolution).
The matrix $\tilde{A} \in \mathcal{R}^{N \times N}$ stores the connectivity between each pair of graph nodes (e.g., $A_{ij}$ represents the connectivity between node $v_{i}$ and $v_j$).
In this work, we measure the node connectivity between $v_{i}$ and $v_{j}$ using the feature similarity between pixel $p_{i}$ and pixel $p_{j}$. 
Particularly, the connectivity between $v_{i}$ and $v_{j}$ can be measured as: $\hat{A}_{ij} = \langle H_{i}, H_{j} \rangle$, where $\langle , \rangle$ is an inner product operation and $H_{i}$ is the $i$-th line in $H$.
Therefore, the feature similarity matrix $\tilde{A}$ can be formulated as:
\begin{equation}
    \tilde{A} = H H^{T}.
\end{equation}

Let $C = \{ \tilde{A}_{ij} | i,j = 1, 2, ..., N \}$ be a set containing all connectivity values in a graph and $b = \{ i | v_{i} \, is \, a \, boundary \, node \}$ be a set including all the indexes of boundary nodes. 
Then, the connectivity related to boundary nodes can be expressed as a set: $C_{b} = \{ \tilde{A}_{ij} | i=b, j=1,2, ..., N\}$. 
To introduce the boundary discrimination to the graph, 
we emphasize the connectivity set $C_{b}$ by re-weighting the feature similarity matrix $\tilde{A}$.
We name this connectivity emphasis as the `boundary-aware operation'. 
Let $B \in \mathcal{R}^{N \times 1}$ be a boundary score map, where $B_{i}$ indicates the probability that $v_{i}$ is a boundary node. 
To maintain the symmetry, the boundary-aware similarity matrix $\tilde{A}_{bw}$ can be formulated as:
\begin{eqnarray}
    \tilde{A}_{bw} &=& (\tilde{A} \odot B + \tilde{A}) + (\tilde{A} \odot B^{T} + \tilde{A}) \nonumber \\
                   &=& \tilde{A} \odot B + \tilde{A} \odot B^{T} + 2\tilde{A},
\end{eqnarray}
where $\odot$ is the matrix dot product, and the terms $\tilde{A}$ added within two brackets are used to maintain the basic semantic similarity among features.

\subsection{Global Reasoning with Efficient Graph Convolution}
\begin{figure*}[tp]
\centering
\includegraphics[width=1\textwidth]{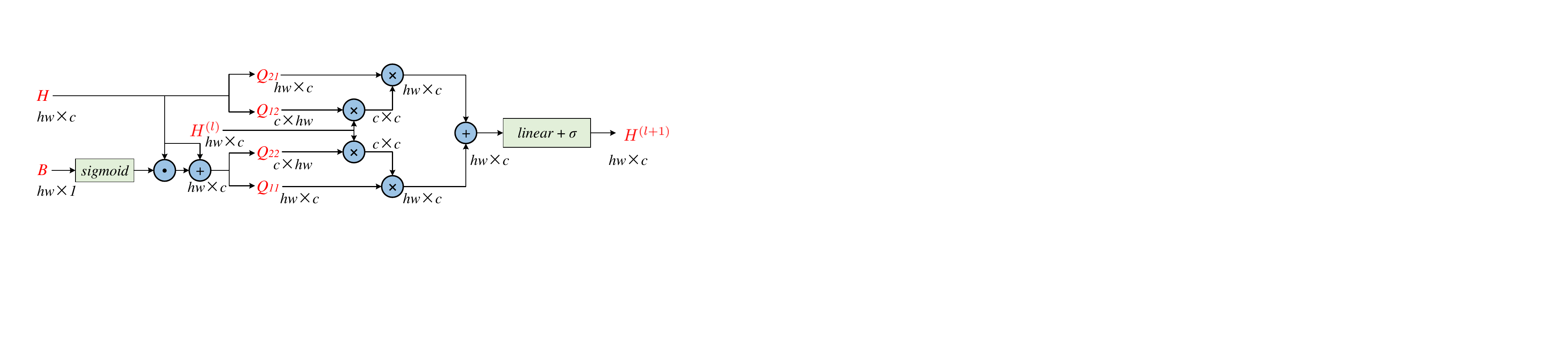}
\caption{Implementation details of graph convolution ($l$-th layer) on the boundary-aware graph with $H^{(l)}$ as input and $H^{(l+1)}$ as output. $H$ is the original node feature matrix, $B$ is the boundary score map, $Q_{11}$, $Q_{12}$, $Q_{21}$, and $Q_{22}$ are decomposed variables of the graph Laplacian, and $\oplus$, $\odot$, and $\oplus$ denote the matrix multiplication, matrix dot product, and matrix summation, respectively.}
\label{fig2}
\end{figure*}

The graph convolution layer deployed in the BGR module can be formulated by rewriting Eq. \ref{eq1} as:
\begin{equation}
    H^{(l+1)} = \sigma (\tilde{D}^{-\frac{1}{2}}_{bw} \tilde{A}_{bw} \tilde{D}^{-\frac{1}{2}}_{bw} H^{(l)} \Theta^{(l)}),
\end{equation}
where the complexity is $O(N^{2})$.
To reduce the computational complexity, we do not explicitly compute $\tilde{A}_{bw}$. 
Alternatively, applying Eq. (3) to Eq. (4), we have:
\begin{eqnarray}
\tilde{A}_{bw} 
               &=& H H^{T} \odot B + HH^{T} + H H^{T} \odot B^{T} +HH^{T} \nonumber\\
               &=& \hat{H} H^{T} + H \hat{H}^{T},
\end{eqnarray}
where $\hat{H}=H \odot B + H$. Hence, as shown in Figure \ref{fig2}, the graph convolution ($l$-th layer) on the boundary-aware graph can be formulated as:
\begin{eqnarray}
H^{(l+1)} &=& \sigma [\tilde{D}^{-\frac{1}{2}}_{bw} \tilde{A}_{bw} \tilde{D}^{-\frac{1}{2}}_{bw} H^{(l)} \Theta^{(l)}]     \nonumber \\
          &=& \sigma [[\tilde{D}^{-\frac{1}{2}}_{bw} \hat{H} H^{T} \tilde{D}^{-\frac{1}{2}}_{bw} H^{(l)}
          + \tilde{D}^{-\frac{1}{2}}_{bw} H \hat{H}^{T} \tilde{D}^{-\frac{1}{2}}_{bw} H^{(l)}]\Theta^{(l)}] \nonumber \\
          &=& \sigma [[Q_{11} (Q_{12} H^{(l)}) 
          + Q_{21} (Q_{22} H^{(l)})] \Theta^{(l)}],
\label{GCN-layer-eq}
\end{eqnarray}
where $Q_{11}=\tilde{D}^{-\frac{1}{2}}_{bw} \hat{H}$, 
$Q_{12}=H^{T} \tilde{D}^{-\frac{1}{2}}_{bw}$, 
$Q_{21}=\tilde{D}^{-\frac{1}{2}}_{bw} H$, 
$Q_{22}=\hat{H}^{T} \tilde{D}^{-\frac{1}{2}}_{bw}$, and
$\Theta^{(l)}$ is trainable parameters in the $l-$th layer to linearly embed node features, and $\sigma[\cdot]$ is the ReLU function.
The input of $l$-th layer, denoted by $H^{(l)}$, is updated to the global context enhanced feature $H^{(l+1)}$.
In addition, $\tilde{D}_{bw}$ can be computed by:
\begin{eqnarray}
\tilde{D}_{bw} = diag[\tilde{A}_{bw} \vec{1}]
               = diag[\hat{H} (H^{T} \vec{1}) 
               + H (\hat{H}^{T} \vec{1})] ,
\label{degree-matrix-eq}
\end{eqnarray}
where $\vec{1} \in \mathcal{R}^{N \times 1}$ is an all-one vector, and $diag[\cdot]$ generates a diagonal matrix by placing $N$ elements at the diagonal of the matrix. 
Note that in Eq. (\ref{GCN-layer-eq}) and (\ref{degree-matrix-eq}), we calculate the terms in the inner brackets firstly. 
In this way, we reduce the complexity from $O(N^{2})$ to $O(Nc)$ and $c << N$.  

\subsection{Embedding BGR Module}
After graph reasoning, the global context enhanced feature is reshaped back to the size of $h\times w\times c$ and summed with the original 2D backbone feature in a residual manner.
Then, the summed feature maps are forwarded to the segmentation backbone for a final dense pixel prediction.
In practice, we simply apply a sequence of $1\times1 Conv+BN+ReLU+1\times1 Conv$ layers to obtain the boundary score map.
Note that the boundary branch only receives the gradient back propagation from the supervision of boundary ground-truth, and the boundary-aware operation of the BGR module requires no gradient during training.
As a result, the proposed BGR module is differentiable so that it can be trained in an end-to-end manner when embedded into existing segmentation backbones at different stages.

\section{Experiments}
To evaluate the proposed method, we conducted comprehensive experiments on three semantic segmentation benchmarks, $i.e.$, the PASCAL VOC dataset \cite{everingham2010pascal}, COCO-Stuff dataset \cite{caesar2018coco}, and Cityscapes dataset \cite{cordts2016cityscapes}.
In this section, we first introduced the datasets and implementation details. 
Then, we gave extensive ablation studies and analyses of our BGR module on PASCAL VOC val set.
Finally, we provided the comparisons against state-of-the-arts (SOTAs) on three benchmarks.

\subsection{Experimental Details}
\textbf{Dataset and Benchmark.} 
The PASCAL VOC dataset is one of the most widely used semantic segmentation datasets which has 10,582 training (augmented by \cite{hariharan2011semantic} with more annotations), 1,449 validation, and 1,456 test images, with the annotations of 20 foreground and 1 background categories. 
The COCO-Stuff dataset provides 9,000 training and 1,000 test images, with the rich annotations of 80 object categories and 91 stuff categories.
The Cityscapes dataset contains 2,975 training, 500 validation, and 1,525 test images, with the high quality annotations of 19 urban scene segmentation categories, and we only used fine annotated data for this study.
For all datasets, we reported the segmentation results in terms of the mean intersection-over-union (IoU).

\textbf{Implementations.}
We adopt ResNet-101 \cite{he2016deep} pretrained on ImageNet \cite{russakovsky2015imagenet}) as the backbone.
Following \cite{chen2018deeplab,fu2019dual,li2020spatial}, we remove the down-sampling operations in last two residual layers of ResNet-101 to set the output stride to 8 and adopt the multi-grid strategy \cite{chen2017rethinking} to enlarge the receptive field.
Before passing backbone features to the BGR module, we first employ the atrous spatial pyramid pooling (ASPP) module \cite{chen2018deeplab,chen2018encoder} to aggregate multi-scale contextual information.
Then, we up-sample the features to the output stride of 4 where the boundary branch and our BGR module are embedded.
We configure two graph convolutional layers in our BGR module and set the feature dimension to 128 throughout the module.
Finally, we apply a linear convolutional layer and an up-sampling operation to the output feature of BGR module to generate a pixel-wise prediction.
We follow \cite{yuan2020segfix} to generate the binary boundary mask upon the down-sampled ground-truth as the supervision of the boundary branch and employ the standard cross-entropy loss and binary cross-entropy loss for main segmentation branch and boundary detection branch, respectively.
The weights of two losses are set to 1 without further tuning.

We implement our method based on Pytorch and conduct all experiments on 4 V100. 
Following prior works \cite{chen2018encoder,zhao2017pyramid}, we employ a poly learning rate with a power of 0.9.
The initial learning rate is set to 0.007, 0.001, and 0.01 for PASCAL VOC, COCO-Stuff, and Cityscapes, respectively.
The momentum and weight decay coefficients are set to 0.9 and 0.0001 respectively.
The training iteration is set to 60K for COCO-Stuff and Cityscapes, and 30K for PASCAL VOC.
We use synchronized batch normalization \cite{zhang2018context} for all experiments and set the batch size to 8 for Cityscapes and 16 for the other datasets.
For data augmentation, we apply random scaling (0.5 to 2), random horizontal flipping, and cropping during training.
The cropping size is set to $513\times513$ for PASCAL VOC and COCO-Stuff, and $769\times769$ for Cityscapes, respectively.

\begin{table*}[tb]
\centering
\setlength{\tabcolsep}{8mm}{
\begin{tabular}{l|c}
\hline
Method                                              & mIoU(\%) \\ \hline \hline
ResNet-101                                          & 77.90    \\
ResNet-101 + ASPP                                   & 79.19    \\
ResNet-101 + ASPP + Boundary Branch                 & 79.28    \\ \hline
ResNet-101 + ASPP + Boundary Branch + BGR(8)        & 80.48    \\ 
ResNet-101 + ASPP + Boundary Branch + BGR(4)        & 80.67    \\ \hline
\end{tabular}}
\caption{Ablation study on PASCAL VOC val set. (8) and (4) represent the output stride of the input feature of the BGR module is 8 and 4, respectively. The segmentation performance is reported in terms of mean IoU ($\%$)}
\label{table1}
\end{table*}

\subsection{Ablation Studies}
In this subsection, we present the ablation studies of our BGR module on the PASCAL VOC val set.
To reduce the training cost and speed up, we conduct all ablation experiments with the output stride of the segmentation backbone being set to 16. 

\textbf{Impact of Network Components.}
To verify the effectiveness of our BGR module, we first conduct ablation study to analyze the contribution of each network component in Table \ref{table1}.
We directly up-sample the features generated by ResNet-101 to the original image resolution as the baseline, which can only obtain an mIoU of 77.90\%.
It shows that, despite enlarging the receptive field, the multi-grid strategy remains far from enough to achieve accurate segmentation. 
Adding the ASPP module to ResNet-101 can improve mIoU by 1.39\%, which indicates the introduced multi-scale feature aggregation is critical for semantic segmentation task.
Besides, we combine ResNet-101 and ASPP module as the backbone and add the boundary detection branch, which can slightly improve mIoU by 0.09\%.
It shows that simply introducing the extra boundary supervision to the segmentation backbone tends to have limited benefit to the segmentation task.
Then, we further embed our BGR module into the network, which substantially achieves a performance gain of 1.2\% in mIoU.
This result demonstrates the effectiveness of BGR module for boosting the segmentation performance in semantic segmentation.
Note that comparing to those approaches which rely on complex boundary detection branches and specifically designed loss functions, the boundary branch with BGR module is quite simple and can be easily embedded into the segmentation backbone.
Finally, we also compare the performance of embedding the BGR module with different output strides.
We separately up-sample the backbone feature maps to the output stride of 8 and 4 for the input of BGR module.
The results in Table \ref{table1} show that using the feature maps with the output stride of 4 can obtain better performance, which can attribute to the feature maps with larger resolution tending to have finer details and boundary information.

\begin{figure*}[tb]
\centering
\includegraphics[width=1\textwidth]{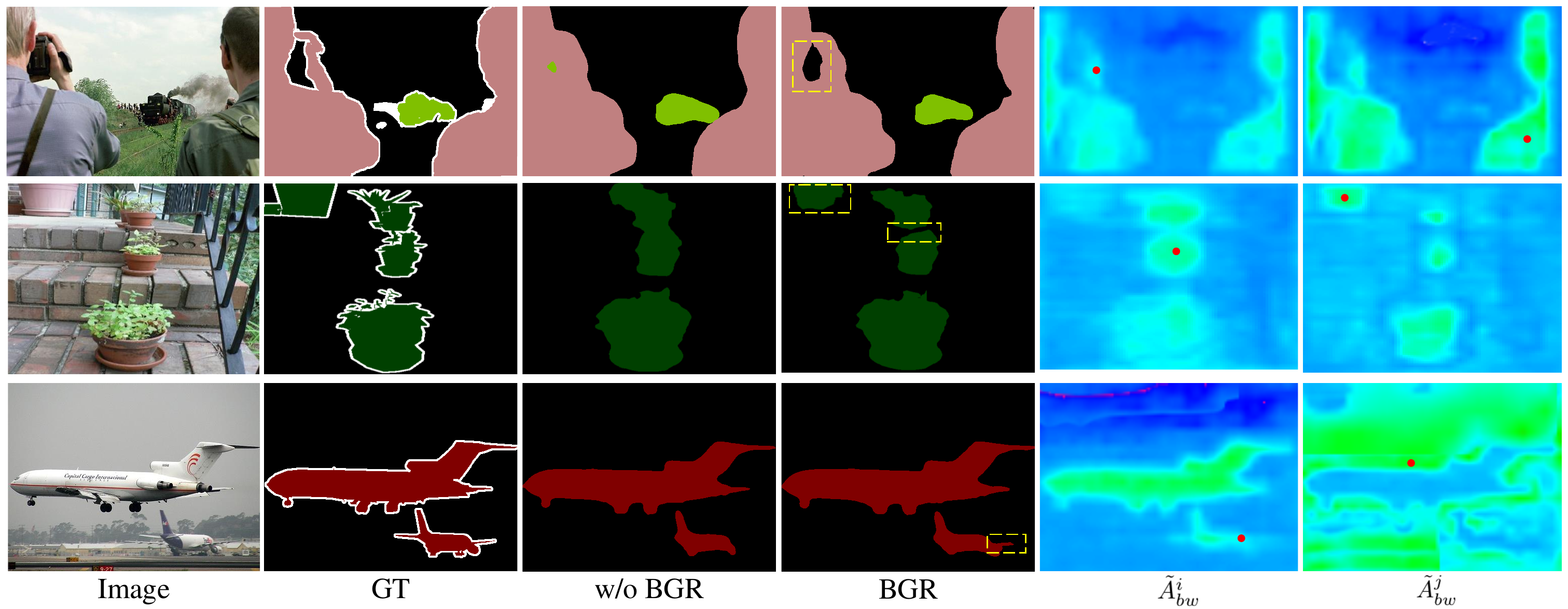}
\caption{Visualization of the results obtained on the PASCAL VOC val set. For each image, we show its ground-truth, segmentation results w/o and with the proposed BGR module. Meanwhile, we give the visualizations of the boundary-aware similarity maps of two selected pixels $i$ and $j$ (marked with red dot).
Better view with colors and zooming in.}
\label{fig3}
\end{figure*}

\begin{table*}[t]
\begin{minipage}[t]{0.35\textwidth}
\centering
\makeatletter\def\@captype{table}\makeatother
\setlength{\tabcolsep}{1mm}{
\begin{tabular}{l|cc|c}
\hline
Method          &MS             & Flip          &mIoU(\%)       \\ \hline \hline
BGR             &               &               &80.48           \\
BGR             &               &\checkmark     &80.88           \\ 
BGR             &\checkmark     &               &81.09           \\
BGR             &\checkmark     &\checkmark     &81.37           \\ 
\hline
\end{tabular}}
\caption{Comparison the performance of different inference strategies on PASCAL VOC val set.}
\label{table2}
\end{minipage}
\hspace{1em}
\begin{minipage}[t]{0.6\textwidth}
\centering
\makeatletter\def\@captype{table}\makeatother
\setlength{\tabcolsep}{1mm}{
\begin{tabular}{l|ccc}
\hline
Method                              &FLOPs (G)      &Memory (M)     &Params (M) \\ \hline \hline
Non-local \cite{wang2018non}        &18.14          &1205           &0.13        \\     
DANet \cite{fu2019dual}             &23.26           &1625           &1.01        \\
GloRe \cite{chen2019graph}          &\textbf{3.49}  &\textbf{1111}           &0.25        \\ \hline
\textbf{BGR}$^{*}$                  &47.74          &1473           &0.10        \\
\textbf{BGR$^{\dag}$}               &6.84           &1141           &\textbf{0.10} \\ \hline
\end{tabular}}
\caption{Complexity comparison of different methods with the feature map of $1\times256\times65\times65$. BGR$^{\dag}$ and BGR$^{*}$ represent deploying graph convolution w and w/o our efficient implementation, respectively.}
\label{table3}
\end{minipage}
\end{table*}

\textbf{Improvement of Different Inference Strategies.}
In the inference phase, we follow \cite{fu2019dual,yu2018learning,li2019expectation,li2020spatial} to adopt multi-scale inference (0.5, 0.75, 1.0, 1.25, 1.5, 1.75) and horizontal flipping for the segmentation of each image.
Here we conduct an ablation study to show the performance gain achieved by different inference strategies.
The backbone features are up-sampled to the output stride of 8 as the input of BGR module for all comparisons.
The experimental results shown in Table \ref{table2} indicate that solely adopting horizontal flipping and multi-scale inference improves the performance by 0.40\% and 0.61\%, respectively, and using both strategies simultaneously can boost the performance by 0.89\% in mIoU. 

\subsection{Visualization of the BGR Module}
In this subsection, we present some visualization results (Figure \ref{fig3}) to show the effectiveness of the BGR module.
The third column shows the segmentation results generated by the backbone (ResNet-101 with ASPP), while the fourth column shows the results with our BGR module employed.
It is clear that using our BGR module can produce consistently better segmentation, especially on some boundary and tiny regions, $e.g.$, the wing of the plane and the gap between the leaves and flowerpot.
In addition, we also provide some visualizations to demonstrate the effectiveness of the BGR module for capturing long-range dependencies.
Concretely, we select two pixels for each image and visualize their corresponding rows in the boundary-aware similarity matrix $\tilde{A}_{bw}$. 
The heatmaps in last two columns show the proposed BGR module can indeed capture long-range dependencies with specific semantic information.
For example, in the first row of $\tilde{A}_{bw}^{i}$, the red dot locates at the hand of the left person, and the heatmap accurately highlights most of the regions of two persons.
Moreover, the flowerpot in the bottom left corner of the second image is difficult to identify since a part of leaves is hidden.
However, the $\tilde{A}_{bw}^{j}$ of a pixel inside this flowerpot can also highlight some parts of other three flowerpots. It indicates that the BGR module can capture, to some extent, the long-range relationship between this inconspicuous flowerpot and the others.
Finally, two heatmaps shown in the last row also highlight the regions which have the same semantic information with selected pixels, even those selected pixels are close to the boundary.

\begin{table*}[tb]
\centering
\setlength{\tabcolsep}{8mm}{
\begin{tabular}{l|c|c|c}
\hline
\multicolumn{1}{l|}{\multirow{2}{*}{Method}} & \multirow{2}{*}{Backbone} & \multicolumn{2}{c}{PASCAL VOC}                     \\ \cline{3-4} 
\multicolumn{1}{c|}{}                        &                           & \multicolumn{1}{c|}{Val} & \multicolumn{1}{c}{Test} \\ \hline \hline
DANet \cite{fu2019dual}             &ResNet-101 &80.40          &82.6      \\
PSPNet \cite{zhao2017pyramid}       &ResNet-101 &-              &82.6      \\
DeepLabV3+ \cite{chen2018encoder}   &ResNet-101 &80.57          &-         \\
DFN \cite{yu2018learning}           &ResNet-101 &80.60          &82.7      \\
ENcNet \cite{zhang2018context}      &ResNet-101 &-              &82.9       \\ 
EMANet \cite{li2019expectation}     &ResNet-101 &80.94          &-          \\ \hline
\textbf{BGR}                        &ResNet-101 &\textbf{81.37} &\textbf{83.8}          \\ \hline
\end{tabular}}
\caption{Comparisons with other state-of-the-art methods on the PASCAL VOC val set and test set.}
\label{voc-test}
\end{table*}

\subsection{Complexity Analysis}
To manifest that our proposed efficient graph convolution implementation strategy can reduce the computational cost of our BGR module, we compare the computational complexity of our BGR using the original graph convolution implementation or using our efficient implementation, shown as BGR$^{*}$ and BGR$^{\dag}$ in Table \ref{table3}, respectively.
Meanwhile, we also provide the computational complexity of other modules, which were proposed for capturing long-range dependencies.
Here we choose two self-attention modules ($i.e.$, Non-local \cite{wang2018non} and DANet \cite{fu2019dual}) and another graph reasoning module (GloRe \cite{chen2019graph}) for comparisons.
For a fair comparison, we set the size of input feature map to $1\times256\times65\times65$ and the inter-channel to 128 for all methods.
For each method, the floating-point operations (FLOPs), GPU memory, and number of parameters are summarized in Table \ref{table3}.
It reveals that our BGR module has fewest parameters and its FLOPs and GPU memory are significantly lower than Non-local and DANet. 
While, the FLOPs and GPU memory of our method are a little bit higher than GloRe since the GloRe projects a group of pixels to a graph node so that the size of constructed graph is much smaller than ours.
In addition, it also shows that using our graph convolution implementation strategy can reduce the GPU memory cost and FLOPs of the BGR module dramatically.
To sum up, our BGR module can perform graph reasoning in a pixel-to-node manner without introducing extra computational overhead.

\subsection{Comparison with State-of-the-art Methods}
\begin{figure*}[tb]
\centering
\includegraphics[width=1\textwidth]{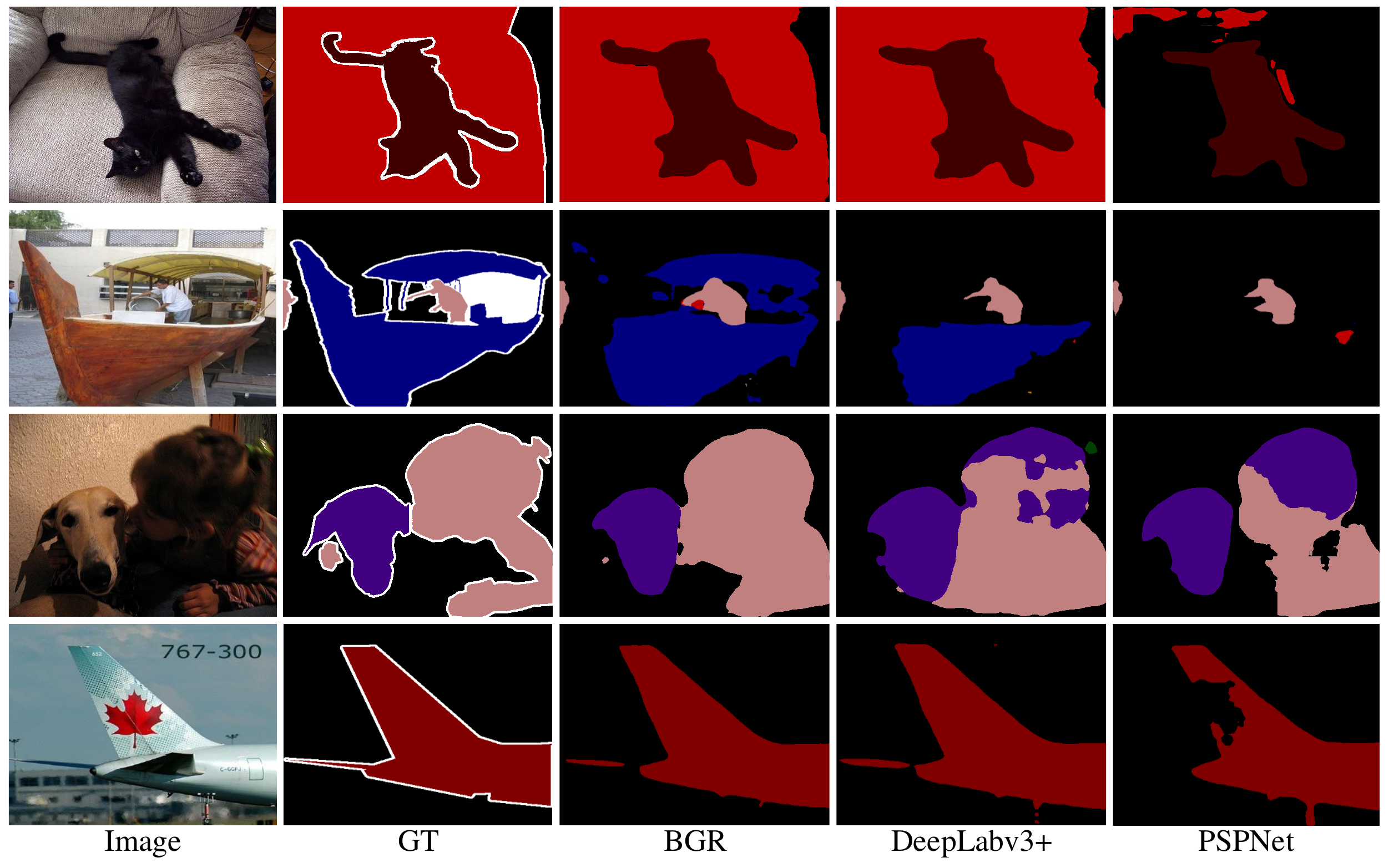}
\caption{Visualization of the results produced by our BGR module and competing methods. The original image, the corresponding ground-truth, as well as the segmentation results of our method, DeepLabv3+ \cite{chen2018encoder}, and PSPNet \cite{zhao2017pyramid} are listed in each row, respectively.}
\label{fig4}
\end{figure*}

To show the superiority of our BGR module, we compare our method with a series of SOTAs on the PASCAL VOC val and test sets, the COCO-Stuff test set, and the Cityscapes test set. 
Our results (Table \ref{voc-test},\ref{coco-test} and \ref{city-test}) suggest that our method with the BGR module achieves the state-of-the-art performance on all three datasets.
Particularly, Table \ref{voc-test} shows the results of different methods on the PASCAL VOC val set and test set.
It shows that our method outperforms DANet, a recent self attention based method for long-range contextual dependency modeling, by $0.97\%$ on val set and $1.2\%$ on test set, with a significantly lower computational cost (see Table \ref{table3}).
In addition, Table \ref{coco-test} and Table \ref{city-test} suggest that our method consistently outperforms previous graph reasoning based methods, $i.e.$, SGR, GloRe, and SpyGR.
Specifically, our BGR module outperforms SGR and GloRe by $1\%$ on the COCO-Stuff test set and the Cityscapes test set, respectively.
Such results demonstrate the superiority of our method, which could be owed to the introduction of boundary-aware information as prior knowledge to graph reasoning, which facilitates the segmentation on boundary regions. 
Some segmentation examples are visualized in Figure \ref{fig4}, which shows that the predictions generated by our method match better with the ground-truth  than those generated by competing methods.

\begin{table}[]
\begin{minipage}[!t]{0.5\textwidth}
\centering
\makeatletter\def\@captype{table}\makeatother
\begin{tabular}{l|l|l}
\hline
Method                              & Backbone  & mIoU(\%)  \\ \hline\hline
CCL \cite{ding2018context}          &ResNet-101 &35.7       \\
DSSPN \cite{liang2018dynamic}       &ResNet-101 &37.3       \\
SGR \cite{liang2018symbolic}        &ResNet-101 &39.1       \\
DANet \cite{fu2019dual}             &ResNet-101 &39.7       \\
CCNet \cite{huang2019ccnet}         &ResNet-101 &39.8       \\
EMANet \cite{li2019expectation}     &ResNet-101 &39.9       \\
SpyGR \cite{li2020spatial}          &ResNet-101 &39.9       \\ \hline
\textbf{BGR}                        &ResNet-101 &\textbf{40.1}      \\ \hline
\end{tabular}
\caption{Comparisons on the COCO-Stuff test set.}
\label{coco-test}
\end{minipage}
\hspace{0.1em}
\begin{minipage}[!t]{0.475\textwidth}
\centering
\makeatletter\def\@captype{table}\makeatother
\begin{tabular}{l|l|l}
\hline
Method                              & Backbone      & mIoU(\%)  \\ \hline\hline
PSPNet \cite{zhao2017pyramid}       &ResNet-101     &78.4       \\
DFN \cite{yu2018learning}           &ResNet-101     &79.3       \\
PSANet \cite{zhao2018psanet}        &ResNet-101     &80.1       \\
DenseASPP \cite{yang2018denseaspp}  &ResNet-101     &80.6       \\
GloRe \cite{chen2019graph}          &ResNet-101     &80.9       \\
DANet \cite{fu2019dual}             &ResNet-101     &81.5       \\
SpyGR \cite{li2020spatial}          &ResNet-101     &81.6       \\ \hline
\textbf{BGR}                        &ResNet-101&\textbf{81.9}   \\ \hline
\end{tabular}
\caption{Comparisons on the Cityscapes test set.}
\label{city-test}
\end{minipage}
\end{table}

\section{Conclusion}
To improve the capability of learning the long-range contextual dependencies for semantic segmentation models, we propose the BGR module which introduces the boundary information as the prior knowledge to the graph reasoning. 
The boundary-aware operation in the BGR module enables boundary-regions to gain more information from the global view thereby facilitates the segmentation on these regions.
The proposed BGR module is compatible to be embedded into existing segmentation backbones and the graph reasoning implementation of BGR module is light-weight with our proposed efficient graph convolution strategy. 
Extensive experiments on three public benchmarks on semantic segmentation clearly demonstrate the effectiveness of our BGR module. 
Hopefully, the proposed BGR module can be enlightening and beneficial for exploring long-range dependencies on semantic segmentation tasks in the computer vision community at large.   

\section*{Broader Impact}
Our approach has shown good capability in exploiting rich feature representation and can be further incorporated into various related tasks.
Therefore, it can facilitate the development of the computer vision community.
However, considering the network configuration and the training strategy of this work are all designed for specific tasks and datasets, one should be discreet to apply our method to some special applications, $e.g.$, auto-driving system or computer-assisted medical intervention.
Directly deploying our method without cautiously fine-tuning on large-scale data related to the corresponding task may lead to unstable predictions and decisions, which may have some negative societal impacts.

\bibliographystyle{IEEEtran}

\end{document}